\documentclass[11pt,a4paper]{article}
\usepackage[hyperref]{acl2020}
\usepackage{times}
\usepackage{latexsym}

\usepackage{microtype}
\usepackage{graphicx}
\aclfinalcopy % Uncomment this line for the final submission
\title{Adversarial NLI for Factual Correctness in Text Summarisation Models}
\author{
	\begin{tabular}{c}
		Mario Barrantes \thanks{equal contribution} \\ 
		\normalfont{\texttt{mariobarque@gmail.com}}
	\end{tabular}% \\
	\begin{tabular}{c}
		Benedikt Herudek \footnotemark[1] \\ 
		\normalfont{\texttt{benedikt.herudek@gmail.com}}
	\end{tabular} \\ 
	\begin{tabular}{c}
		\textbf{Richard Wang} \footnotemark[1] \\
		\normalfont{\texttt{richcmwang@gmail.com}}
	\end{tabular}
}

\date{}

\begin{document}
\maketitle
\begin{abstract}
We apply the Adversarial NLI dataset to train the NLI model and show that the model has the potential to enhance factual correctness in abstract summarization.  We follow the work of \citet{falke-etal-2019-ranking}, which rank multiple generated summaries based on the entailment probabilities between an source document and summaries and select the summary that has the highest entailment probability.  The authors' earlier study concluded that current NLI models are not sufficiently accurate for the ranking task.   We show that the Transformer models fine-tuned on the new dataset achieve significantly higher accuracy and have the potential of selecting a coherent summary. 
\end{abstract}

\section{Introduction}
Text summarization aims at compressing long textual documents into a short form that contains essential information from the source.  Recently pre-trained Transformer Architectures significantly improve text summarization and achieve high benchmark scores \cite{liu-lapata-2019-text}.  A commonly used metric to measure the quality of the summary is ROUGE, which stands for Recall-Oriented Understudy for Gisting Evaluation \citep{lin-2004-rouge}. Rouge works by comparing the overlapping subsequence between an automatically produced summary against a set of reference summaries.  A substantial criticism of ROUGE is that it allows summaries that contradict to the original text but still achieve a high score.   Multiple studies \citep{kryscinski-etal-2019-neural}, \citep{falke-etal-2019-ranking} show that current SOTA abstractive summarization models generate more than a third of incorrect summaries, either producing new information or contradicting to the source document.  \citet{kryscinski-etal-2019-neural} pointed out that ROUGE score is uncorrelated to human evaluation, and does not ensure factual correctness.  Another shortcoming of ROUGE is that it excludes semantically similar words between the reference summary and the generated summary.  A summary of similar meaning to the reference summary may have low ROUGE score because it uses different words.

Natural Language Inference (NLI) models have been incorporated into summarization model for improving factual correctness through multi-task learning \citep{guo-etal-2018-soft} or selecting the generated summary \citep{falke-etal-2019-ranking} that has the highest entailment probability. Our investigation follows the work of the later authors.  Their method generates several abstractive summaries from the current state-of-art-method \citep{chen-bansal-2018-fast}, ranks them with the entailment probability between the source document and each generated summary, and selects the summary that has the highest probability. The authors found that current NLI models cannot perform well on such a task. Using a dataset created for investigating the applicability of NLI models to their ranking task,  the authors show that in more than 34\% of the time all current NLI models suggest a summary not entailed by the source document.  They conclude that improving the factual correctness of abstractive summarization requires a new NLI model.

Studies have shown that current NLI models do not genuinely learn to reason, but rather exploit spurious statistical patterns \citet{gururangan-etal-2018-annotation} \citet{Poliak:2018}. \citet{Geirho:2020} pointed out that such "shortcut learning" is collective in deep learning models.  Recently \citet{nie2019adversarial} developed the Adversarial NLI dataset (ANLI) that aimed to improve these issues. The authors employ an adversarial process between human and model to construct increasingly challenging examples, pushing the model to learn correct reasoning. We hypothesis that the ANLI trained models would be more robust, and therefore more likely to produce a more accurate entailment probability suitable for selecting a coherent summary. We fine-tune pre-trained transformer models with the ANLI dataset, the resulting models will outperform the existing NLI Models in selecting the coherent summaries described in \citep{falke-etal-2019-ranking}.  To validate our hypothesis, we test the fine-tune classifiers on the validation set provided by the authors and compare our results with the authors' results shown in the paper.

\section{Related Work}

Many works in summarization use the Transformer-type of model architecture. \citet{JZhang:2019} propose a pre-trained Transformer-based encoder-decoder model leveraging a massive dataset. \citet{liu2018generating} propose a hybrid method that uses tf-idf for content selection and memory-efficient encoder-only Transformers to process lengthy sequences to generate Wikipedia articles.  \citet{Yoon:2020} trains on semantic similarity between the reference summary and the generated summary with pre-trained BART and a Semantic Similarity layer instead of using cross-entropy for a loss function. 

"Traditional" sequence-to-sequence attention type of models exist in the literature for summarization task. \citet{cao2017faithful} focus on sentence-level summarization. The work uses two attention mechanisms to incorporate OpenIE or dependency parsing to ensure the consistency of the summarized sentence to the original sentence. \citet{falke-etal-2019-ranking} build their work on top of a hybrid summarization model \citep{chen-bansal-2018-fast} and rank several generated summaries with NLI models. \citet{guo-etal-2018-soft} propose a multi-task learning strategy that learns summarization, question generation and entailment generation simultaneously. 

Entailment relation has been used to verify or enhance the correct logics between a source document and a summary. \citet{guo-etal-2018-soft} proposed multi-task learning entailment generation to ensure the summary logically follows the source documents. \citet{falke-etal-2019-ranking} studied whether entailment using models trained on current datasets such as SNLI and MNLI could effectively detect errors in the generated summaries. They found that out-of-the-box NLI models do not perform well on detecting the incorrect fact conditioned on the source documents. They claim current state-of-art models trained on SNLI \citep{bowman-etal-2015-large} and MNLI \citep{Williams:2018} rely on heuristics and do not transfer well to newswire text of CNN-DM and more diverse genres are needed. 

\citet{kryscinski-etal-2019-neural} provide critiques on current status of abstract summarization. The authors point out that in current settings models are given a source document with one associated reference summary and no additional information, leaving the task of summarization under constrained and too ambiguous to be solved by end-to-end models. \citet{guo-etal-2018-soft} proposed a Multi-task model that includes question generation, aiming to guide the constrained summary to answer target questions. \citet{Yoon:2020} identify the same problem, but present it in a different angle and provide an opposite solution.  The two different perspectives seem to be two sides of the same coin.  \citet{kryscinski-etal-2019-neural} believe that a summary problem should be constrained to be solved effectively. \citet{Yoon:2020} adopts an opposite thought: since the summary is under-constrained, many equally valid summaries may co-exist. Therefore, current cross-entropy loss calculated at a token level for a gold reference is too limited and needs to be changed. \citet{Yoon:2020} propose using semantic similarity as a training goal with an impressive human evaluation to support their hypothesis.  

The critics also point out the evaluation problem. ROUGE score is based on exact match and does not support synonymous phrases. The observation also enhances the argument made by \citep{Yoon:2020} about the rigidity of cross-entropy loss trained at a token level, which does not support synonymous phrases. The studies in the critics show that up to 30\% of summaries generated by abstractive models contain factual inconsistencies. Most of the models only provide evidence of model performance through ROUGE and human evaluation and do not presents the factual accuracy through direct examination similar to the one presented in Falke et al. 2019. It remains unclear how effective these models \citep{liu-lapata-2019-text}, \citep{cao2017faithful}, \citep{guo-etal-2018-soft}, \citep{Yoon:2020} in the factual accuracy aspect.

Contrast to \citep{falke-etal-2019-ranking}, in subsequent work, \citet{Kryscinski2019EvaluatingTF}  argued that checking factual consistency on a sentence-sentence level is insufficient and proposed a document-sentence approach for factual consistency checking. Their method uses an automatic evaluation method that generates training data by applying rule-based transformations to the source documents and training a BERT based model for verifying factual consistency. The evaluation method improves over models using MNLI.  

The critics also point out the layout bias in current training data. Many of the state-of-the-art summarization models are trained on the news dataset such as CNN/DM. Because news articles adhere to an "Inverted Pyramid" writing structure, the authors show that models' performance is strongly affected by the layout bias of the news data. \citet{Poliak:2018} pointed out similar problems encountered for the NLI models, which exploit spurious statistical patterns in the training data to reach high F1 scores. The new dataset ANLI \citet{nie2019adversarial} address NLI issues. Summarization communities would also benefit from a new dataset that avoids the bias.  

Many state-of-the-art summarization models use pre-trained Transformer types of models and achieve high ROUGE scores. However, only very limited number of works perform direct examination to verify factual correctness. ROUGE does not correlate well with the quality of a summary performed by human evaluation, providing little information about the coherency of a summary.  Current NLI models do not offer a reliable tool for automatic factual checking or enhancing. ANLI dataset deliberately addresses the weakness of the current NLI dataset and has potential to improves error detection in summarization.

\section{Data}
We use Pre-trained Transformer Language Models based on the hugging face API. The datasets used for pretraining these Language Models can be found in the model references.  

For fine-tuning the models, we begin with MNLI dataset \citet{Williams:2018} downloaded from the tensorflow dataset website.  Our main dataset for fine-tuning the Language Models is ANLI dataset by \citet{nie2019adversarial}.   ANLI aims to address the weakness \citet{gururangan-etal-2018-annotation} \citet{Poliak:2018} found in earlier dataset such as SNLI and MNLI. Table 2 shows a summary of ANLI dataset.  It consists of 162k training examples, 2.2k dev examples, and 2.2k testing examples. The source of the data mainly comes from Wiki, and are generated in three increasingly challenging rounds through an adversarial process between human and model.  We believe models trained on the challenging ANLI dataset have great potential to improve the performance of the ranking task of summarization in \citet{falke-etal-2019-ranking}

For the test set, we use a summary correctness dataset (SC) created by \citet{falke-etal-2019-ranking}.   The dataset includes 373 triples that consists of a source sentence, an incorrect summary sentence and a correct summary sentence. For each triple, we create two pairs; one is a source sentence d and an incorrect summary $s_{-}$, the other pair has  the same source sentence d, but with a correct summary $s_{+}$.  We then compute the entailment probabilities $N(d, s_{-})$, $N(d, s_{+})$ for each pair to determine how well a model performs.  The detailed performance measure is defined in next section.

\begin{table}
	\includegraphics[width=\linewidth]{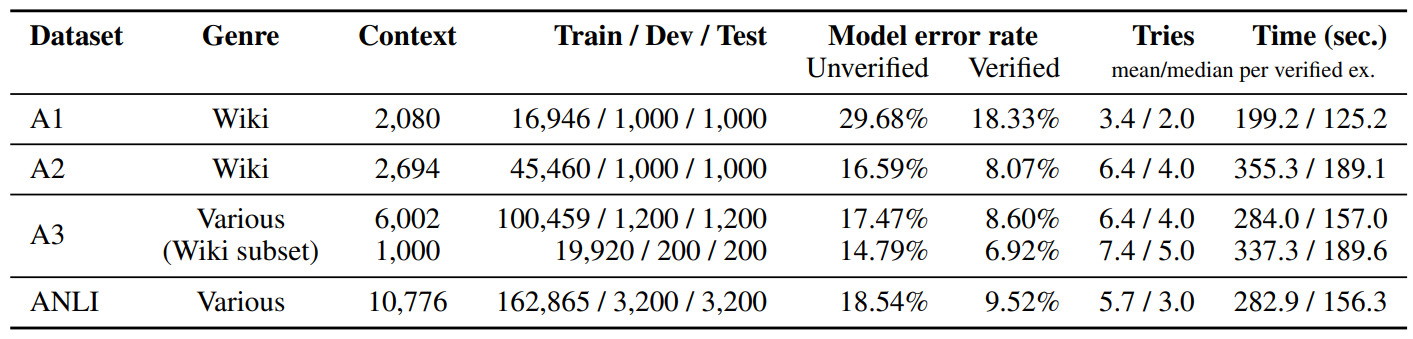}
	\caption{ANLI Dataset Summary}
	\label{fig:anlidataset}
\end{table}

\section{Model}

We use pretrained transformer models and fine-tune on MNLI and ANLI dataset for our experiment. We begin with BERT-Base \citep{devlin-etal-2019-bert}, and then use more powerful RoBERTa-Base \citep{DBLP:journals/corr/abs-1907-11692} and XLNet-Base \citep{Yang_1:2019}.

\section{Experiment}

\paragraph{Evaluation Metrics}
We use simple accuracy as the metric to measure the performance of the model fine-tuned on the ANLI dataset. As described in the Dataset section, each example triple in the SC dataset creates two pairs:  a source sentence d and an incorrect summary $s_{-}$ pair, and the same source d and a correct summary $s_{+}$ pair.   We compute the entailment probability $N (d, s_{-})$ and $N (d, s_{+})$  for each pair, and the frequency of how often a NLI model prefers the wrong summary, i.e. $N (d, s_{-}) \geq  N (d, s_{+})$.  The relative frequency of such incorrect order will give an indication of the NLI model performance for the summary ranking task.  The lower the incorrect frequency, the higher the accuracy.  In short, the accuracy is defined as

\begin{equation}
\frac{ \#\{d:N(d,s_{-}) < N(d,s_{+}) \} }{\# \textit{ of examples}}
\end{equation}We use the performance of the BERT based model trained on MNLI as the benchmark.  Falked et al. reported that around 36\% of the test examples in SC dataset are incorrect, which yields to 64\% of accuracy.   We will compute the accuracy for the ANLI tuned models and see whether they have higher accuracy on the SC test set.
We note that for general summarization tasks, because of no other effective measurement, ROUGE scores is still the prevailing metric to measure the performance of a summarization model.  Higher performance of a NLI model for the summary ranking task may also improve the ROUGE score.

\paragraph{Model and Baseline}
For our experiments, we generally follow the guidance from the ANLI paper \citet{nie2019adversarial}.  We train NLI models with three transformer-type architectures, including BERT \citep{devlin-etal-2019-bert}, RoBERTa \citep{DBLP:journals/corr/abs-1907-11692} and XLNet \citep{Yang_1:2019}.  \citet{nie2019adversarial} trained the large version of these models with SNLI and MNLI as a base level, and then added FEVER and ANLI. Their results show that the performance of the models generally improves as more datasets are included in the fine-tuning (table 3).  Due to the time and resource constraints,  we will only use the base version of these models and train them only on MNLI and ANLI.  We experiment the results for 1) only MNLI and 2) both MNLI and ANLI.  We then see the effects of ANLI datasets on the model performance.

Our baseline comes from the results presented in \citep{falke-etal-2019-ranking}. The authors presented the frequency of incorrect orders, and we adjusted the results to the percentage of correct orders, see Table \ref{accuracy_ranking_paper}.  We will compare our results against their BERT result that attains 64\% accuracy.

\begin{table}
	\centering
	\begin{tabular}{lrl}
		\hline \textbf{NLI Model} & \textbf{Accuracy} \\ \hline
		Random & 50.0\% \\
		\hline 
		DA & 57.4\% \\
		InferSent & 58.7\% \\
		SSE & 62.7\% \\
		BERT & 64.1\% \\
		ESIM & 67.6\%\\
		\hline
	\end{tabular}
	\caption{\label{accuracy_ranking_paper} Accuracy from \citet{falke-etal-2019-ranking} }
\end{table}

\paragraph{Training}
We use Pre-trained Transformer Language Models BERT-base, RoBERTa-base an XLNet-base with classification task from Hugging Face \footnote{While using Hugging Face, we used a 128 lenth due to time and computational ressources}. We note that due to the resource limitation, we restrict input length to 128. Some of the training examples are truncated.  The incomplete context and hypothesis in the input may have some impact on the performance of the training results of the NLI model. We reuse their GLUE script and add the functionality to train new ANLI dataset.  We first fine-tune each model with MNLI dataset and compute the accuracy.  Then we further train each of the resulting model with ANLI dataset.

\paragraph{Results}

We replicated the experiment by downloading the dataset about summary correctness, where we have 373 contexts, and for each context there is an incorrect hypothesis and an correct hypothesis.  Here a context represents the source document, and a hypothesis represents a summary.Our evaluation computes the proportion of the examples that have higher entailment probability in the correct context- hypothesis pair than the incorrect one as defined in the Metrics section. Table \ref{accuracy result} shows our accuracy results. We see model fine-tuned on ANLI improve about 4\% with BERT and RoBERTa, and has only 1\% improvement with XLNet. We also notice that our BERT model tuned on MNLI achieves about 7\% more accuracy  (71\% versus 64\%) than reported in \citet{falke-etal-2019-ranking}.  We do not know the reason from the information on paper but presume it could be due to different implementation.  The highest achiever is RoBERTa with MNLI + ANLI dataset, which scores almost 15\% higher than the best results obtained usng ESIM model reported by the authors, see Table \ref{accuracy_ranking_paper}.

\begin{table}
	\centering
	\begin{tabular}{lrl}
		\hline \textbf{Dataset} & \textbf{NLI Model} & \textbf{Accuracy} \\ \hline
		 & BERT & 71.31\% \\
		MNLI & RoBERTa & 78.55\% \\
		 & XLNet & 79.35\% \\
		 \hline 
		 & BERT & 75.33\% \\

		MNLI + ANLI & RoBERTa & 82.57\% \\
		 & XLNet & 80.96\%\\
		\hline
	\end{tabular}
	\caption{\label{accuracy result} Accuracy results }
\end{table}

\section{Analysis}

\begin{figure}
	\includegraphics[width=\linewidth]{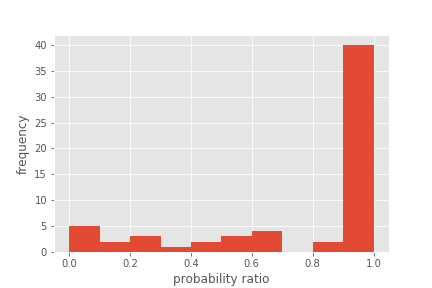}
	\caption{The entailment probability ratio $\frac{N(d,s_{+})}{N(d,s_{-})}$
	} of a correct summary and an incorrect summary.
	\label{fig:stats}
\end{figure}

To understand better about the ANLI trained model,  we analyze the incorrect selection, i.e. $N(d,s_{-}) > N(d,s_{+})$ with RoBERTa. Figure \ref{fig:stats} shows a frequency distribution of the entailment probability ratio between a correct summary and an incorrect summary.  Around 66\% the probability ratio is greater than 0.9, meaning the correct and incorrect entailment probability are close. On the other end, five examples have probability ratio less then 0.1, indicating that the model strongly prefers a summary that is not entailed by the context document.  We examine three of them below. We list each example in the order of a context sentence, a correct summary and an incorrect summary. 

In the first example, the model strongly prefers the sentence without \emph{the}.  It could be related to the fact that the source sentence has no \emph{the} in front of the departments:

\begin{quote}
	
"...... mr sarkozy said: 'never in the history of the fifth republic has our political family won so many departments.",

"sarkozy's electoral alliance has won \emph{the} departmental elections.",

"sarkozy's electoral alliance has won departmental elections."	

\end{quote}

In the next example, the model confuses about the two different persons and takes them as the same person:

\begin{quote}
"the man, named only as saleh, said the masked militant who appears in several beheading videos was a senior figure in the extremist organisation responsible for murdering foreign captives.",

"the man, named only as saleh, said the masked militant appears in several beheading videos.",

"the man, named only as saleh, appears in several beheading videos."
\end{quote}

Finally, "the three" was taken as a noun such as the three boys, but in fact the source document means three-year-old.  The model could also prefers keeping the same past perfect tense: 

\begin{quote}
	"we reported that, just before we battened down the hatches, all the three and four-year-old boys had been snapping at their pyjama elastic and comparing the size of their willies.",
	
	"the three and four-year-old boys were snapping at their pyjama elastic.",
	
	"the three and boys had been snapping at their pyjama elastic."

\end{quote}

Our ANLI trained models apparently are not ready to handle these subtle cases.

\section{Discussion}

We have shown that NLI models trained on ANLI dataset improve accuracy for selecting the factual correctness summmary.  While our experiment supports the hypothesis, we are interested in understanding how ANLI dataset \citet{nie2019adversarial} imapcts the transformer models through observation of attention coefficients.   Various studies have presented analysis of attention mechanism.  \citet{clark2019does} present studies on various aspect of linguistic notion to attention coefficients for BERT model. \citet{DBLP:journals/corr/abs-1902-10186} show a different perspective that the correlation between learned attention weights and measures such as accuracy for classification is weak.  We attempt to understand the trained model through the attention using a visualization tool \citet{Vig:2019}.

\begin{figure}
	\includegraphics[width=\linewidth]{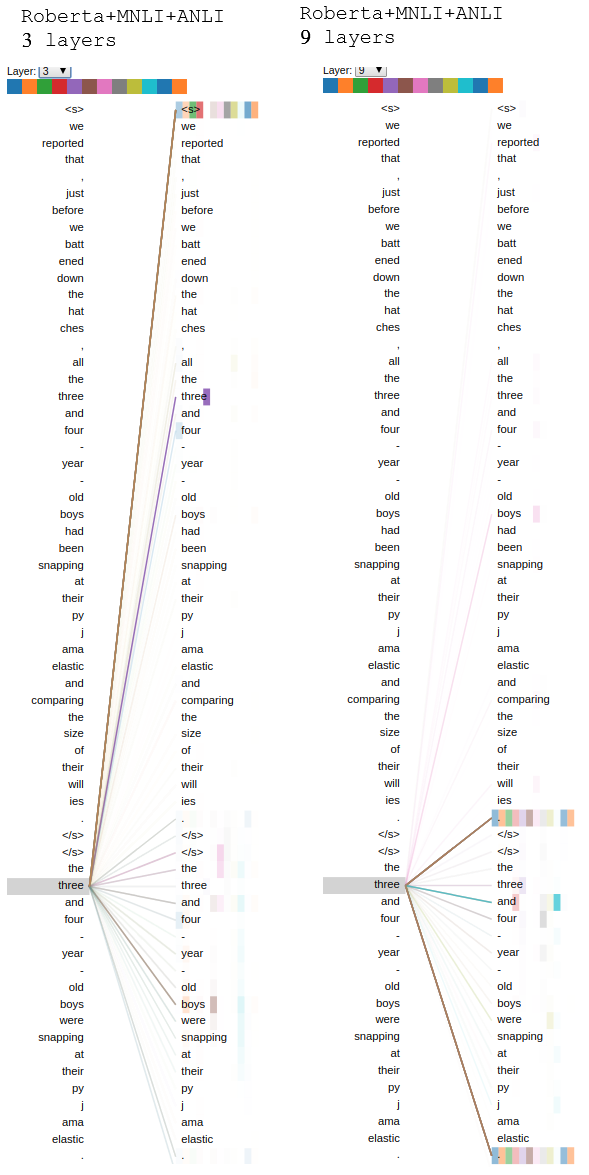}
	\caption{An example of a document and its correct summary pairs of the self attention head view from a RoBERTa model with MNLI + ANLI datasets at layer 3 at the left and layer 9 at the right.  The attentions go across the context and hypothesis.
	}
	\label{fig:correct}
\end{figure}

Figure \ref{fig:correct} shows the observation of attentions using the correct summary of the example in previous section.  We notice earlier layers attend across a context and its hypothesis whereas later layers attend within a context or a hypothesis.  This seems to suggest the relation between a context and a hypothesis is largely encoded in the early layers. We also notice in figure \ref{fig:correct}, the \emph{three} in the hypotheis allocates no weights to \emph{year-old} in the context, which could contribute to the fact that \emph{three} is regards as a separate noun and unrelated to \emph{year-old}.

We compared the impact of ANLI dataset by observing attention coefficients of Roberta + MNLI vs Roberta + MNLI + ANLI models. We picked an example where the second model has a significant probability improvement:

\begin{quote}
	"sex addiction is described by the relationship counselling service relate as any sexual activity that feels out of control ...",
	
	"sex service relate to sexual activity that feels out of control.",
	
	"sex addiction is also described by the relationship counselling service relate."
\end{quote}Again, the first sentence is the original document, and the second and third ones are correct and incorrect summaries, respectively.

\begin{figure}
	\includegraphics[width=\linewidth]{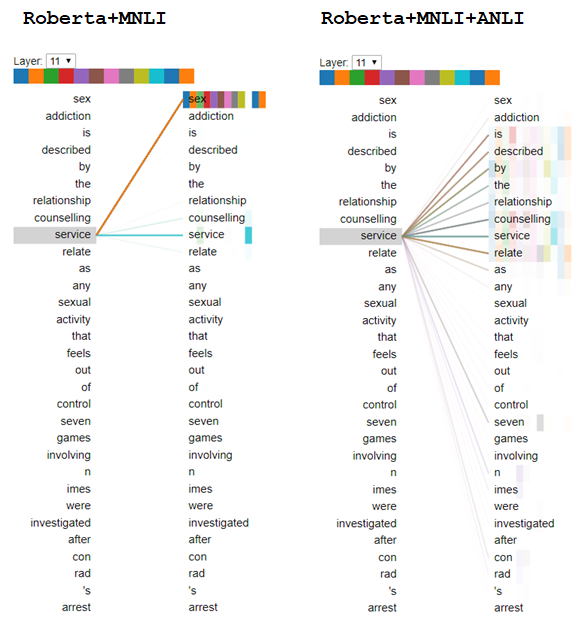}
	\caption{(left) the  attention  headview visualization from  a  RoBERTa  model  fine  tuned  with  MNLI and (right) the  same  view for  RoBERTa  model  fine  tuned  with  MNLI  and  ANLI. Both visualizations shows context and hypothesis.
	}
	\label{fig:visual}
\end{figure}

Figure \ref{fig:visual} shows the context of the attention head view for both models highlighting the token service. For Roberta + MNLI the token \emph{service} has a strong attention to the token \emph{sex}, even though they do not have proximity in the context. Once we add the ANLI we see that strong attention relation fades away and now service has more attention weights to other tokens such as \emph{counselling} or \emph{relate}. We assume that since the first model tokens \emph{sex} and \emph{service} have strong attention relationship, the model choose the incorrect summary.

\paragraph{A Note on Transformers Interpretability}
Studies have shown that Neural Network often can be significantly reduced while predicting essentially the same results.  For example, \cite{NIPS1989_250} used 'Optimal Brain Damage' and allows more than half of the weights from a neural network while keeping the performance of the network stable.   Much research also exists in reducing the size of transformers in NLP community.   Distillation methods  \cite{Jiao_1:2019} train a smaller 'student model' to simulate the capacity of the larger 'teacher model.  \cite{DBLP:journals/corr/abs-1905-10650} proposed pruning the entire model such as removing weights, attention heads or the whole layers.  We believe a reduced transformer model allows researchers to focus on essential parts, opening the opportunity for better understanding and interpreting, for example, how it classifies a relation correctly.  There is a body of research about interpreting neural networks \cite{2017arXiv170208608D}, \cite{Molnar:2019}, \cite{2018arXiv180201933G}.  In particular, \cite{Narang:2020} point out that humans usage of languages cannot easily obtain a full and transparent view of their decision-making process. We rely on other humans to explain our judgments in language. The authors train a text-to-text model performing on different NLP tasks to explain the predictions. Just as humans formulate explanations for human behavior, neural networks should have the opportunity to explain their predictions using natural text.

%How exactly ANLI changes the transformer models and lead  attention relation is an open question for future work. One assumption is that since large Transformers such as RoBERTa encodes current human knowledge from large datasets, tokens that appear together frequently on those datasets have strong attention, which work most of the time, but sometimes it defies the real meaning of a sentence. And since ANLI methodology was challenging humans to find those soft spots in current state of the art models, it could have an impact on the encoding.

\section{Conclusion}
In this paper we showed Transformer Models fine tuned on a new Adversarial Natural Language Inference Dataset (ANLI) improves factual correctness. The new model can be used for abstract summarization tasks via a ranking approach presented in \citep{falke-etal-2019-ranking}. We follow the work \citet{nie2019adversarial} to fine-tune BERT, RoBERTa and XLNet models with MNLI and ANLI dataset and test its performance for selecting correct summary on a dataset provided by \citet{falke-etal-2019-ranking}.  Our quantitative Analysis demonstrates signficantly higher accuracy in MNLI + ANLI trained NLI models compared to the results in \citep{falke-etal-2019-ranking}. We attempt to gain understanding on the model through the attention mechanisms with the visualization tool provided in \citep{Vig:2019}. 

\paragraph{Future Work}
We believe using a bigger dataset than the one we used from \citep{falke-etal-2019-ranking} can validate further our work and provide new insights. Also a more fine grained analysis could be done through the different rounds of ANLI dataset \citet{nie2019adversarial}. 
Finally, expanding our experiment to fine tune smaller and faster transformers like DistilBERT \citet{Sanh2019DistilBERTAD} will provide important data when inference time is a concern.

\section*{Acknowledgments}

We thank Prof Christopher Potts and the teaching team for their support and guidance during the course XCS224U at Stanford Center for Professional Development. We thank Yixin Nie for the discussion on model tuning.

\bibliography{acl2020}
\bibliographystyle{acl_natbib}

\end{document}